\pdfoutput=1

\documentclass[11pt]{article}

\usepackage[preprint]{acl}

\usepackage{times}
\usepackage{latexsym}

\usepackage[T1]{fontenc}

\usepackage[utf8]{inputenc}

\usepackage{microtype}

\usepackage{inconsolata}

\usepackage{graphicx}

\usepackage{multirow}
\usepackage{booktabs}
\usepackage{amsmath}
\usepackage{amssymb}
\usepackage[ruled,vlined]{algorithm2e}
\usepackage{algpseudocode}
\usepackage{xcolor}
\usepackage{bm}
\usepackage{makecell}

\title{Explainable Depression Detection in Clinical Interviews with Personalized Retrieval-Augmented Generation}

\author{Linhai Zhang$^{1}$,
        Ziyang Gao$^{2}$,
        Deyu Zhou$^{2}$,
        Yulan He$^{1,3}$ \\
        $^1$King's College London \hspace{0.5cm}
        $^2$Southeast University  \hspace{0.5cm}
        $^3$The Alan Turing Institute \\
        \texttt{\{linhai.zhang, yulan.he\}@kcl.ac.uk} \\
        \texttt{\{ziyanggao, d.zhou\}@seu.edu.cn}
  }

\begin{document}
\maketitle
\begin{abstract}
Depression is a widespread mental health disorder, and clinical interviews are the gold standard for assessment. 
However, their reliance on scarce professionals highlights the need for automated detection. 
Current systems mainly employ black-box neural networks, which lack interpretability, which is crucial in mental health contexts. 
Some attempts to improve interpretability use post-hoc LLM generation but suffer from hallucination.
To address these limitations, we propose RED, a Retrieval-augmented generation framework for Explainable depression Detection. 
RED retrieves evidence from clinical interview transcripts, providing explanations for predictions. 
Traditional query-based retrieval systems use a one-size-fits-all approach, which may not be optimal for depression detection, as user backgrounds and situations vary. 
We introduce a personalized query generation module that combines standard queries with user-specific background inferred by LLMs, tailoring retrieval to individual contexts.
Additionally, to enhance LLM performance in social intelligence, we augment LLMs by retrieving relevant knowledge from a social intelligence datastore using an event-centric retriever. 
Experimental results on the real-world benchmark demonstrate RED's effectiveness compared to neural networks and LLM-based baselines.
\end{abstract}

\section{Introduction}

Depression is one of the most prevalent mental health disorders, affecting millions of individuals worldwide~\cite{fava2000major}. 
Timely detection is crucial for effective intervention, yet traditional assessment methods, such as clinical interviews, rely heavily on trained professionals, which are in short supply~\cite{kroenke2001phq}. 
As a result, there is an increasing need for automated systems capable of accurately detecting depression from patient interactions, enabling faster and more widespread diagnosis~\cite{islam2018depression, orabi2018deep, chen-etal-2024-depression}.

\begin{figure}[t]
    \centering
    \includegraphics[width=0.49\textwidth]{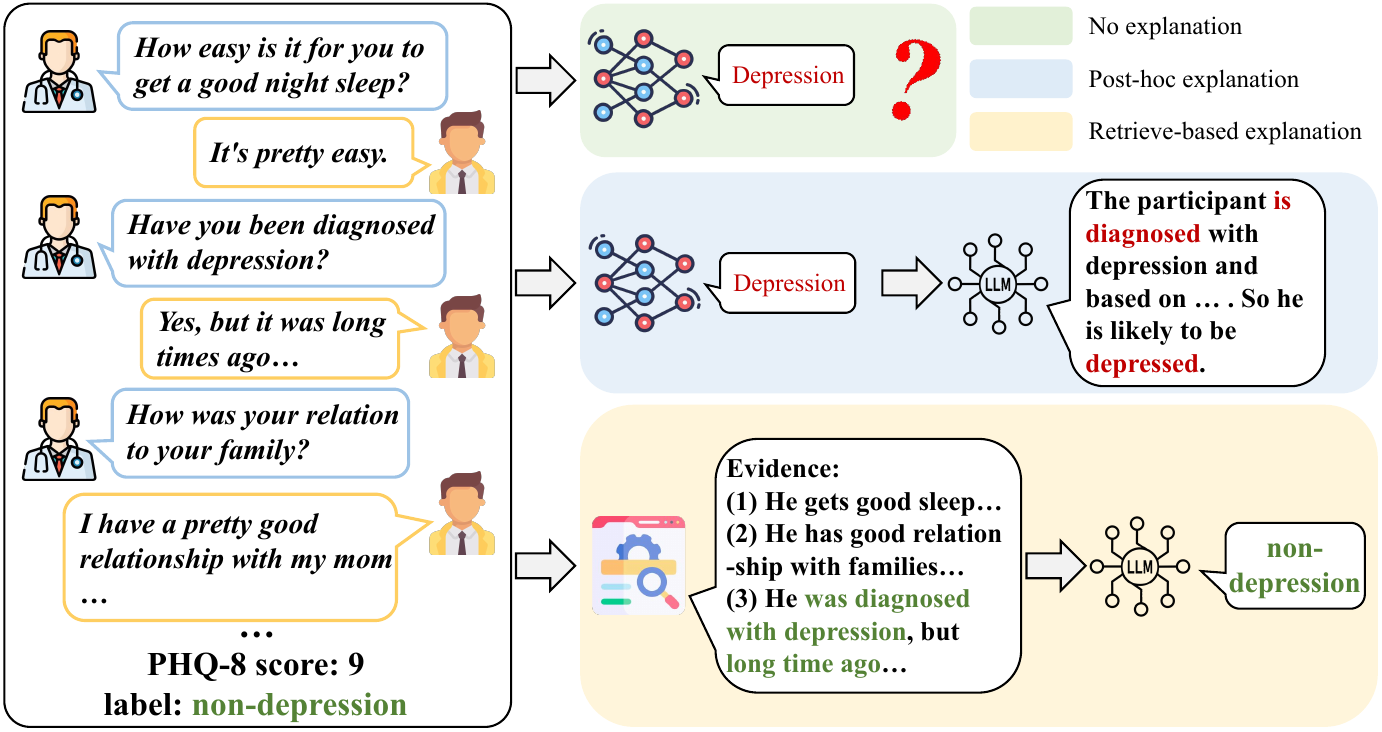}
    \caption{Comparison between different depression detection methods. Most of the methods focus on improving performance while ignoring the explanation. Some work tries to generate post-hoc explanations with LLMs while suffering from the hallucination. Our work employs a RAG-based framework to retrieve the supporting evidence from dialogue, which serves as the explanations for the predictions.}
    \label{fig:intro}
\end{figure}

Most automated depression detection methods currently focus on enhancing system performance using various approaches. Previous work has concentrated on aggregating word representations for prediction~\cite{mallol2019hierarchical, burdisso2023node}, or further incorporating affective and mental health lexicons~\cite{xezonaki2020affective, villatoro2021approximating}. 
Some studies have explored modalities beyond text, such as audio~\cite{ma2016depaudionet, sardari2022audio}, or have integrated multimodal data~\cite{al2018detecting, wu2022climate}.
However, in health-related tasks, precision is not the sole priority; it is also crucial to understand the rationale behind the system's predictions to make the system more transparent and reliable. 
To achieve this, some studies have utilized the internal states of neural models, such as attention scores~\cite{zogan2022explainable}, though these explanations often fall short of human interpretability.
In response, some researchers have leveraged the generative capabilities of large language models (LLMs) to create post-hoc explanations based on the system’s predictions~\cite{wang-etal-2024-explainable}.

As shown in Figure~\ref{fig:intro}, most previous work overlooks the interpretability of the system, making its predictions difficult to understand and less reliable~\cite{mallol2019hierarchical, burdisso2023node}. Some studies have attempted to generate post-hoc explanations using large language models (LLMs), but these approaches are often plagued by the issue of hallucinated generation~\cite{wang-etal-2024-explainable}.
To address these challenges, we propose employing the Retrieval-Augmented Generation (RAG) framework for explainable depression detection. The RAG framework combines a retriever model with LLMs to improve the LLM's ability to handle content beyond its input window and to update its knowledge~\cite{lewis2020retrieval, gao2023retrieval}.
With the RAG framework, crucial information from the interview dialogue is retrieved and serves as supporting evidence for the LLM's prediction. Furthermore, the retrieval process helps filter out noisy or irrelevant information that could negatively impact the prediction. Since the evidence is directly retrieved from the interview text, it is both human-understandable and free from hallucinations.

In this paper, we propose RED, a Retrieval-augmented generation framework for Explainable depression Detection. 
RED retrieves relevant evidence from clinical interview transcripts and uses this information as an explanation for its predictions. 
This retrieval-based approach ensures that the explanations are grounded in the actual content of the interview, enhancing transparency.
Since depression detection is a personalized task where participants’ backgrounds can vary, the traditional approach of using a single, unified query for all users may lead to suboptimal results.
To address this, we introduce a personal query generation module in RED, which tailors the basic query to each individual based on their profile, inferred from the LLM. 
This customization enables more accurate and context-sensitive predictions.
While LLMs have proven effective across many tasks due to their extensive world knowledge, they often lack domain-specific knowledge and fall short of social intelligence~\cite{wang2023emotional,hou2024entering,liu-etal-2024-interintent}. 
To mitigate this, we enhance the social intelligence of LLMs by retrieving external knowledge from a social intelligence knowledge base using an event-centric retriever.
Experimental results on a real-world depression detection benchmark demonstrate that RED outperforms both neural network-based and LLM-based methods, highlighting its effectiveness.

The contributions of this paper can be summarized as: 
\begin{itemize}
    \item \textbf{New framework}: We propose to perform explainable derepression detection based RAG framework with personal retrieve process;
    \item \textbf{LLM social intelligence enhancement}: A novel module that enhances the social intelligence of LLM based on event-centric retrieval is proposed;
    \item \textbf{Empirical Performance}: Experimental results demonstrate the effectiveness of our approach compared to both neural network-based and LLM-based baselines.
\end{itemize}

\section{Related Work}

\subsection{Depreesion Detection}
Depression detection is challenging due to its subtle nature, with traditional methods relying on clinical interviews or social media data~\cite{gratch2014distress,burdisso2020tau,salas2022detecting}. 
Recent approaches focus on multi-modal data from interviews, combining text, audio, and video for better accuracy~\cite{gratch2014distress, burdisso2020tau, salas2022detecting}. 
These methods aggregate features at various levels (word or utterance) to capture more nuanced signs of depression.
Early risk detection is also gaining traction, using techniques like incremental classifiers and risk window-based methods to predict depression before full symptoms emerge~\cite{burdisso2019text, burdisso2019unsl,sadeque2018measuring}. 
These approaches enable timely interventions by detecting consistent patterns over time.
Recent advancements also incorporate LLMs, which are fine-tuned on mental health datasets to capture complex linguistic and psychological cues. 
By combining LLMs with multimodal data, these methods show promise in improving both depression detection and early intervention~\cite{an2020multimodal,yoon2022d}.

\begin{figure*}[t]
    \centering
    \includegraphics[width=0.9\textwidth]{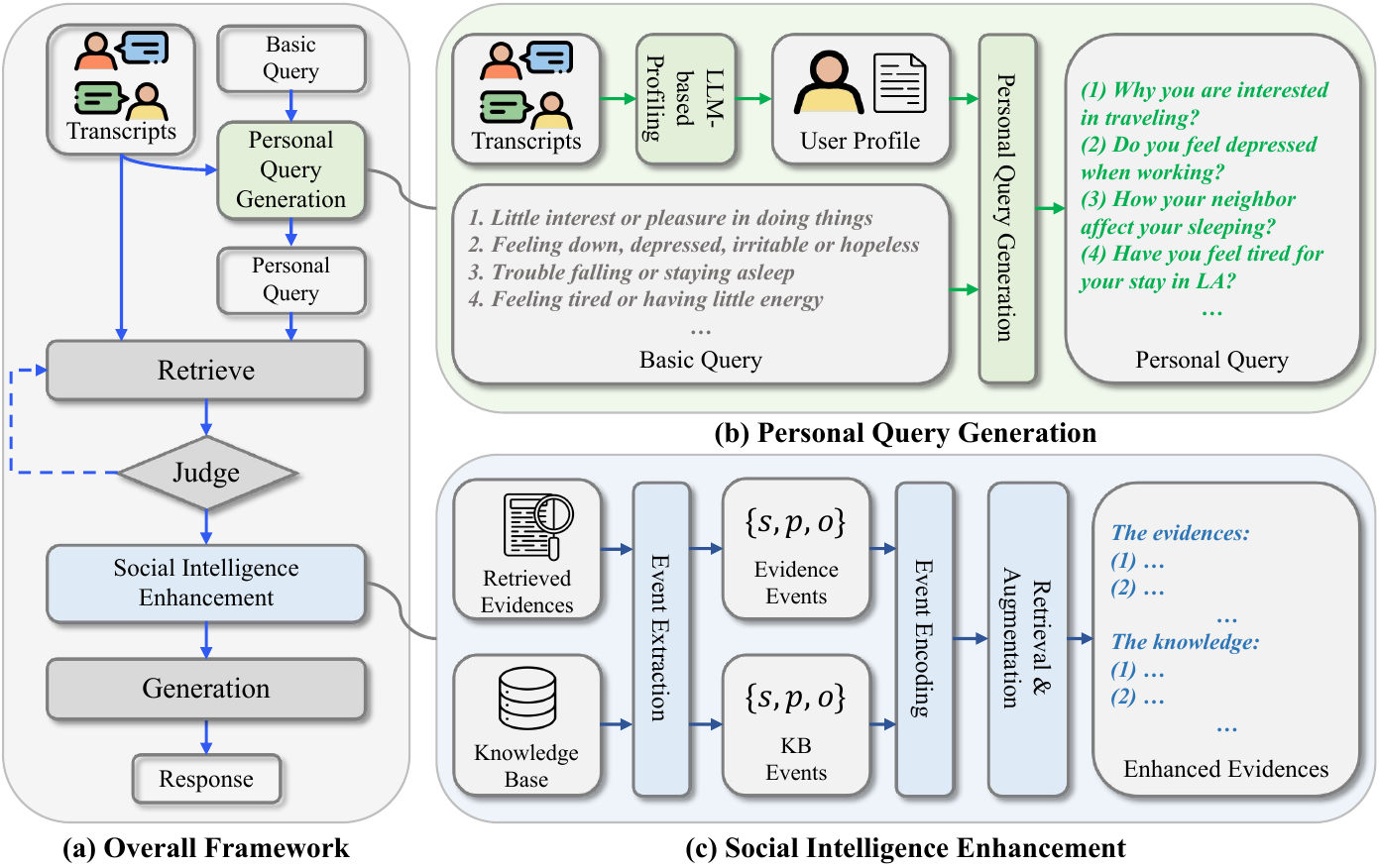}
    \caption{Overview of RED, which consists of (a) The adaptive RAG framework with two important modules, (b) the Personal Query Generation module, and (c) the Social Intelligence Enhancement module.}
    \label{fig:method}
\end{figure*}

\subsection{Retrievel Augementated Generation}
Retrieval-Augmented Generation (RAG) enhances language models (LMs) by incorporating retrieved text passages into the input, leading to significant improvements in knowledge-intensive tasks~\cite{guu2020retrieval,lewis2020retrieval}. 
Recent advancements in RAG techniques have focused on instruction-tuning LMs with a fixed number of retrieved passages or jointly pre-training a retriever and LM followed by few-shot fine-tuning~\cite{luo2023sail,izacard2022few}. 
Some approaches adaptively retrieve passages during generation~\cite{jiang-etal-2023-active}, while others, like \citet{schick2023toolformer}, train LMs to generate API calls for named entities. However, these improvements often come with trade-offs in runtime efficiency, robustness, and contextual relevance~\cite{mallen-etal-2023-trust,shi2023large}. 
To address these challenges, recent work introduces methods like SELF-RAG, which enables on-demand retrieval and filters out irrelevant passages through self-reflection, enhancing robustness and control~\cite{lin2024radit, yoran2024making}. 
SELF-RAG~\cite{asai2023self} also evaluates the factuality and quality of the generated output without relying on external models during inference, making it more efficient and customizable. 
Additionally, other concurrent RAG methods, such as LATS~\cite{zhou2023language}, explore ways to improve retrieval for specific tasks like question answering through tree search.

\section{Method}

\subsection{Overall Framework}
As shown in Figure~\ref{fig:method}, RED utilizes an adaptive RAG framework for depression detection, using retrieved chunks as explanations. 
First, the personal query generation module customizes the basic query based on the inferred user profile. 
Then, the system retrieves relevant evidence for depression prediction based on the personalized query and transcription, while the judgment module produces a stop signal. 
The retrieved evidence is further enhanced with knowledge from the social intelligence knowledge base through an additional retrieval process. 
Finally, the LLM generates the response using the enhanced evidence.

We will discuss the RAG framework, the personal query generation module, and the social intelligence enhancement module in detail.

\subsection{Explainable Depression Detection with Adaptive RAG}

\begin{algorithm}[th]
\caption{RED inference process}\label{alg:red}
\small
\KwData{Dialogue $D$, PHQ-8 aspect set $A=\{a\}$, Basic query set $Q=\{q_a\}$, Knowledge base $K$}
\KwResult{the predicted label $\hat{y}$}
$re=\text{TRUE}$  \\
$Q^{(u)} = \text{PQ-Gen}(Q,D)$ \Comment{\textcolor{green}{Section}~\ref{sec:pqg}} \\
\For{$a \in A$}{
    $D_a = D$ \\
    \While{$re=\text{TRUE}$}{
        $e = \text{Retrieve}(q^{(u)}_a,D_a)$ \Comment{\textcolor{gray}{Retrieve}} \\
        $E_a = E_a \cup e$  \\
        $D_a = D_a / e$  \\
        $re = \text{Judge}(q^{(u)}_a,E_a)$ \Comment{\textcolor{gray}{Judge}} \\
    }
}
$E=\bigcup_{a \in A} E_a$ \\
$E_s = \text{SI-Enh}(E,K)$ \Comment{\textcolor{blue}{Section}~\ref{sec:sie}} \\
$\hat{y}=\text{Geneartion}(E_s)$ \Comment{\textcolor{gray}{Geneartion}} \\
\end{algorithm}

Previous approaches use LLMs to generate post-hoc explanations for depression detection, which often suffer from hallucination issues. 
In this paper, we use Retrieval Augmented Generation (RAG) to first retrieve relevant dialogue snippets from the dialogue set as evidence, then integrate this evidence into the LLM prompts for predictions. 
This approach ensures the system uses reliable, relevant snippets, avoiding irrelevant content and improving precision. 
Additionally, the retrieved snippets serve as explanations for the system's output, enhancing interpretability.
As evidence requirements may vary across users, we employ an adaptive RAG framework that allows the system to determine when to stop retrieving automatically.

The detailed inference process of RED is shown in Algorithm~\ref{alg:red}. 
The inputs include the dialogue $D$, the PHQ-8 aspect set $A={a}$, the basic query set $Q=\{q_a\}$, and the knowledge base $K$. 
The output is the predicted depression label $\hat{y}={0,1}$.
First, RED tailors the basic query $Q=\{q_a\}$ into a personal query $Q^{(u)} = \text{PQ-Gen}(Q,D)$ using the personal query generation module, $\text{PQ-Gen}(Q,D)$, detailed in Section~\ref{sec:pqg}. 
Next, the personal query $Q^{(u)}$ is used to retrieve relevant evidence $e$ from $D$, based on the aspects $A={a}$ aligned with the PHQ-8 questionnaire. 
Then, the social intelligence enhancement module, $\text{SI-Enh}(E,K)$, augments the evidence $E$ with knowledge from the knowledge base $K$, as explained in Section~\ref{sec:sie}. 
Finally, the LLM generates the final response $\hat{y}$ based on the enhanced evidence $E_s$.

\paragraph{Retrieve}
The standard process for depression detection is based on the PHQ-8 (Patient Health Questionnaire-8)~\cite{kroenke2001phq}, where the participant self-evaluates on 8 questions addressing various aspects of depression, such as interest in activities and issues with movement or speech. 
Each question is scored from 0 (Not at all) to 3 (Nearly every day), resulting in a total score from 0 to 24. 
A score above 10 typically indicates depression. 

Building on this process, we propose retrieving evidence based on different aspects. 
For each aspect $a$, a personal query $q_a^{(u)}$ is used to retrieve evidence $e$ from the dialogue set $D_a$, forming the evidence set $E_a$. 
RAG typically uses sparse retrievers (e.g., BM25) and dense retrievers.
In this work, we implement a dense retriever based on GPT embedding model~\footnote{https://platform.openai.com/docs/guides/embeddings} with L2 similarity. 
\begin{equation}
\begin{split}
    \bm{q} &= \text{BERT}(q) \\ 
    \bm{d}_i &= \text{BERT}(d_i) \\
    e &= \text{Top-1}(\{sim(\bm{q},\bm{d}_i)\})
\end{split}
\end{equation}
where $q$ denotes the query, $d_i \in D_a$ denotes the $i$-th dialogue snippet, $sim(\cdot,\cdot)$ denotes the cosine similarity. 
At each iteration, the top-1 result is returned, and both the evidence and dialogue sets are updated by $E_a = E_a \cup e$ and $D_a = D_a / e$.

\paragraph{Judge}
Since dialogue can shift to different topics, a one-time retrieval may be insufficient for judgment. Moreover, determining a specific threshold for retrieval is challenging. 
To address this, we propose a judgment module that allows the system to determine when to stop retrieving adaptively.
The judgment model is a binary classification model, which can be implemented as either a supervised neural network or an LLM agent. 
In this paper, we implement it as an LLM agent that takes the retrieved evidence set $E_a$ and the personal query $q_a^{(u)}$ as inputs:
\begin{equation}
    re = \text{LLM}(q_a^{(u)},E_a|p_j)
\end{equation}
where $re$ is the retrieval indicator and $p_j$ is the judgment prompt. The full prompt can be found in Appendix~\ref{app:prompt}. 

\paragraph{Generation}
The final response is generated using the enhanced evidence set $E_s$:
\begin{equation}
    \hat{y} = \text{LLM}(E_s|p_g)
\end{equation}
where $p_g$ is the depression detection task prompt. The full prompt is provided in Appendix~\ref{app:prompt}.

\subsection{Personal Query Generation}
\label{sec:pqg}
The depression diagnosis interview process is highly personalized and can vary across users~\cite{goldman1999awareness}. 
Thus, using a single query for all participants may lead to suboptimal results.
To address this, we propose tailoring the basic query to each participant’s background, creating a personal query.
Inspired by profile-augmented generation in personalized LLMs~\cite{richardson2023integrating}, we first infer the user profile from dialogues with an LLM agent for participant $u$:
\begin{equation}
    d_u = \text{LLM}(D|p_d)
\end{equation}
where $p_d$ is the user profiling prompt, with the full prompt available in Appendix~\ref{app:prompt}. 
Next, the personal query $q^{(u)}_a$ for aspect $a$ is generated from the basic query $q_a$:
\begin{equation}
    q^{(u)}_a = \text{LLM}(q_a,d_u|q_p)
\end{equation}
where $q_p$ denotes the personal query generation prompt, and the full prompt can be found in Appendix~\ref{app:prompt} along with the basic queries template.

\subsection{Soical Inerllegence Enhancement}
\label{sec:sie}

Although LLMs perform well on various tasks with extensive knowledge, they still lack social intelligence and psychological understanding. 
To enhance LLM judgment, we propose retrieving relevant knowledge from a psychological knowledge base to augment the evidence~\cite{wu-etal-2024-coke}.

However, retrieving relevant knowledge from dialogue can be challenging due to its rich and noisy content. 
Inspired by event-centric sentiment analysis~\cite{zhou-etal-2021-implicit}, we treat key elements in the dialogue as events that happened to the participant. 
This leads to an event-centric retrieval approach. 
By extracting key events from the dialogue snippets, we can focus on relevant information for more accurate retrieval.

To extract events from text, one could use a supervised event extraction model or an LLM agent. 
In this paper, we employ the LLM agent to extract event triplets ${s, p, o}$ from text $t$, where $s$ is the subject, $p$ is the predicate and $o$ is the object:
\begin{equation}
    \{s,p,o\} = \text{LLM}(t|p_e)
\end{equation}
where $p_e$ is the event extraction prompt. We perform event extraction for both the dialogue sentences and the keys in the knowledge base to ensure alignment.

With the extracted events, we perform event representation learning using the event encoder from MORE-CL~\cite{zhang-etal-2023-multi}, where event triplets are projected into a Gaussian embedding space, and similarity is calculated using symmetric KL-divergence. Formally, the knowledge base is represented as $K = {(k_i, v_i)}$, where $k$ represents the key and $v$ represents the value.
\begin{equation}
\begin{split}
    (\bm{\mu},\bm{\sigma}) &= \text{E-encoder}(\{s,p,o\}) \\ 
    \{s\} &= \text{Top-k}(\text{KL}(g_i,g_j)+\text{KL}(g_j,g_i))
\end{split}
\end{equation}
where $\{s\}$ are retrieved knowledge pairs. 

\section{Experimental Settings}

\subsection{Datasets} 
We conduct experiments on an available corpus for clinical depression detection: the Distress Analysis Interview Corpus-Wizard of Oz (DAIC-WoZ)~\cite{gratch2014distress}, which is a widely utilized English-language dataset comprising interviews from 189 participants, with data available in the form of transcripts, audio recordings, and videos. After the interaction, participants are asked to complete the PHQ-8 questionnaire~\cite{kroenke2009phq}, which assesses eight specific symptoms related to depression. These symptoms include loss of interest, feelings of depression, sleep disturbances, fatigue, loss of appetite, feelings of failure, lack of concentration, and reduced movement. Participants scoring 10 or higher are classified as depressed, while those with scores below 10 are classified as control. Detailed statistic of the dataset can be found in Appendix~\ref{app:dataset}. 
Following the prior research~\cite{chen-etal-2024-depression} and the specificity of our methodology, both the development and training sets are utilized for evaluation, as the labels for the test set are unavailable. 

We do not employ another interview-based depression detection dataset EATD~\cite{shen2022automatic} because it is not fully the clinical setting, where each participant was only asked three questions, making the dialogue content too short for retrieval.

For the social intelligence enhancement module, we employ COKE, a cognitive knowledge graph for machine theory of mind~\cite{wu-etal-2024-coke}. COKE contains a series of cognitive chains to describe human mental activities and behavioral/affective responses in social situations. In RED, we concat \textit{situation} and \textit{clue} in COKE as the query for retrieval, and the rest of the elements as values. Detailed for the dataset can be found in Appendix~\ref{app:dataset}. 

\subsection{Implementation Details} 
For the implementation of RED, we employ the GPT model to generate personalized queries based on the basic query and the User Profile, which is summarized from the transcripts using \texttt{gpt-4o-2024-08-06}. In the retrieval phase, we use the \texttt{text-embedding-3-large} embeddings to encode both queries and transcripts and apply the L2 distance metric to retrieve the top \(K\) evidence (with \(K = 10\) as the default). The COKE dataset, which contains rich and diverse scenes, is used as the Knowledge Base. To extract event triplets \(s, p, o\) from text, we utilize a LLM agent. Each event triplet is then encoded into a Gaussian embedding space (\(dims = 500\)) using MORE-CL. The similarity between event triplets is calculated using the L2 distance metric, and the top \(M\) triplets are retrieved (with \(M = 2\) as the default). 

We followed previous studies~\cite{burdisso2023node} and, in addition to considering the depressed, control, and macro F1 scores, we also included precision and recall for both the depressed and control groups, resulting in a total of seven evaluation metrics. We select the checkpoint for evaluation based on macro F1 scores. The final results for comparison are the average scores of 3 runs. We run all experiments on a single NVIDIA GeForce RTX 3090 in Windows 11. For the LLM, we use \texttt{gpt-4o,gpt-4o-mini\allowbreak,gpt-4}, which are provided by OpenAI.  The hyperparameter ranges can be found in Appendix~\ref{app:hyperparameter}.

\subsection{Baselines} 
To validate the effectiveness of the proposed RED for the Depression Detection task, we implement
and compare NN-based methods and LLM-based methods.
\paragraph{NN-based method} 
\textbf{$\omega$-GCN} \cite{burdisso2023node} is an approach for weighting self-connecting edges in a Graph Convolutional Network (GCN) 
\textbf{EATD-Fusion} \cite{9746569} is a bi-modal model that utilizes both speech characteristics and linguistic content from participants' interviews. 
\textbf{MFM-Att}\cite{FANG2023104561} is a multimodal fusion model with a multi-level attention mechanism (MFM-Att) for depression detection, aiming to effectively extract depression-related features.
\textbf{HCAG} \cite{9413486} is a hierarchical Context-Aware Graph Attention Model model that utilizes the Graph Attention Network (GAT) to capture relational contextual information from both text and audio modalities.
\textbf{SEGA} \cite{chen-etal-2024-depression}transforms clinical interviews into a directed acyclic graph and enhances it with principle-guided data augmentation using large language models (LLMs) 
\paragraph{LLM-based method}
\textbf{Direct Prompt} is a prompt-learning method designed to guide large language models (LLMs) in making judgments about depression. 
\textbf{Naive RAG} is a technique that integrates the Retrieval-Augmented Generation (RAG) framework with LLMs. It uses a retriever to search for relevant evidence from a knowledge base or dataset, which is then fed into an LLM to make judgments or generate appropriate responses.
\textbf{Personal RAG} builds upon the previous method by enhancing the query generation process, which is now based on the user profile, ensuring more personalized and contextually relevant evidence retrieval.

The details for the implementation can be found in Appendix~\ref{app:hyperparameter}.

\section{Experimental Analysis}

In this section, we present comprehensive experiments conducted on LaMP. 
Through an in-depth analysis of the results, we aim to address the following Research Questions (\textbf{RQs}):
\begin{itemize}
    \item \textbf{RQ1}: How does RED perform compared to baseline models in a standard setting?
    \item \textbf{RQ2}: What impact do different architectural structures and components have on model performance?
    \item \textbf{RQ3}: How effective RED is in terms of explanation extraction?
    \item \textbf{RQ4}: How does RED perform in qualitative evaluations?
\end{itemize}

\begin{table}[t]
\centering
\resizebox{0.95\columnwidth}{!}{
\begin{tabular}{clccc}
\hline
\multicolumn{2}{c}{Method} & \multicolumn{1}{c}{Depressed} & \multicolumn{1}{c}{Control} & \multicolumn{1}{c}{Marco} \\ \hline
         & $\omega$-GCN    & 78.26   & 89.36           & 83.81                 \\
         & EATD-Fusion     & 69.57   & 85.11           & 77.34                 \\
NN       & MFM-Att         & 78.57   & 85.71           & 82.14                 \\
-based   & HCAG            & 76.92   & 86.36           & 81.64                  \\
         & SEGA            & 81.48   & 88.37           & 84.93                     \\
         & SEGA++          & 84.62   & 90.91           & 87.76         \\ \hline
         & Direct Prompt   & 74.07   & 83.72           & 78.90                   \\
LLM      & Naive RAG       & 78.97   & 88.05           & 84.39                        \\
-based   & Personal RAG    & 79.87   & 88.92           & 84.39                \\
         & RED             & \textbf{87.83} & \textbf{92.17}   & \textbf{90.00}           \\ \hline
\end{tabular}}
\caption{Performance of RED and other baselines on the development set of DAIC-WoZ benchmark. The best scores are in \textbf{bold}. All LLM-based results are an average of three rounds of experiments based on GPT-4o.}
\label{tab:dev}
\end{table}

\begin{table*}[t]
\centering
\resizebox{1.7\columnwidth}{!}{
\begin{tabular}{lccccccc}
\hline
\multicolumn{1}{c}{\multirow{2}{*}{Method}} & \multicolumn{3}{c}{Depressed}                                                        & \multicolumn{3}{c}{Control}                                                          & \multicolumn{1}{c}{Marco} \\
\multicolumn{1}{c}{}                        & \multicolumn{1}{c}{Precision} & \multicolumn{1}{c}{Recall} & \multicolumn{1}{c}{F-1} & \multicolumn{1}{c}{Precision} & \multicolumn{1}{c}{Recall} & \multicolumn{1}{c}{F-1} & \multicolumn{1}{c}{F-1}   \\ \hline
Direct Prompt (GPT-4)      &55.24 &\textbf{79.36} &65.14 &89.40 &73.00 &80.37 &72.75  \\
Direct Prompt (GPT-4o-mini)&57.77 &73.81 &\textbf{73.81} &87.55 &77.33 &82.12 &73.47\\
Direct Prompt (GPT-4o)     &59.64 &78.57 &67.81 &\textbf{89.62} &77.67 &83.21 &75.51 \\  \hline
Naive RAG (GPT-4)          &65.04 &73.81 &69.15 &88.34 &83.33 &85.76 &77.46   \\
Naive RAG (GPT-4o-mini)    &61.64 &76.19 &68.11	&88.89 &80.00 &84.20 &76.16  \\
Naive RAG (GPT-4o)         &68.15 &73.02 &70.49 &88.32 &85.67 &86.97 &78.73 \\ \hline
Peronal RAG (GPT-4)        &\textbf{69.93} &68.26 &69.07 &86.8  &\textbf{87.00} &87.23 &78.15 \\
Peronal RAG (GPT-4o-mini)  &60.66 &69.05 &64.45 &86.21 &81.00 &83.48 &73.96 \\
Peronal RAG (GPT-4o)       &68.98 &72.22 &70.56 &88.09 &86.33 &\textbf{87.20}&\textbf{78.88}\\ \hline

\end{tabular}}
\caption{Performance of RED's variants on the full set of DAIC-WoZ benchmark. The best scores are in \textbf{bold}. All LLM-based results are an average of three rounds of experiments.}
\label{tab:full}
\end{table*}

\subsection{Main Results}

To answer \textbf{RQ1}, we compare the performance of RED with baseline models on the DAIC-WoZ development set. The results, shown in Table~\ref{tab:dev}, demonstrate that RED outperforms all baselines. We observe the following:

\paragraph{RED outperforms all baselines, especially for the depressed class.} 
NN-based methods generally outperform LLM-based baselines with direct prompting, but RED improves LLM performance by incorporating personal retrieval and social intelligence, raising the macro F1 score from 78.90\% to 90.00\%. While F1 scores for the control class are high across all methods, indicating a tendency to classify most participants as control, RED achieves a significant gain in the depressed class by retrieving relevant evidence and enhancing the LLM with the necessary knowledge for accurate predictions.

\paragraph{The retrieval process is still necessary, even if the contents do not exceed the input window size.} 
As shown in Table~\ref{tab:dev}, the LLM baseline underperforms compared to NN-based methods. However, using a naive retrieval process, the LLM-based baseline improves, with macro F1 rising from 78.90\% to 84.39\%, while personal RAG achieves further gains. This improvement is due to the retrieval process filtering out irrelevant information, allowing the model to focus on what matters. Thus, the retrieval process remains essential, even when content doesn't exceed the input window size.

\paragraph{Social intelligence enhancement with calibration brings significant improvement.} 
Personal retrieval improves performance over direct prompting but ties with the non-data-augmented SEGA~\cite{chen-etal-2024-depression}. With the social intelligence enhancement, RED generates fine-grained depression scores for each PHQ-8 category, allowing for calibration. Since DAIC-WoZ transcripts do not cover all PHQ-8 aspects, such as appetite and movement difficulties, RED’s predicted scores are generally lower than the actual scores. By combining social intelligence enhancement with calibration, and adjusting the threshold from 10 to 8, RED achieves substantial improvement, particularly in predicting the depressed class.

\begin{figure*}[t]
    \centering
    \includegraphics[width=0.9\textwidth]{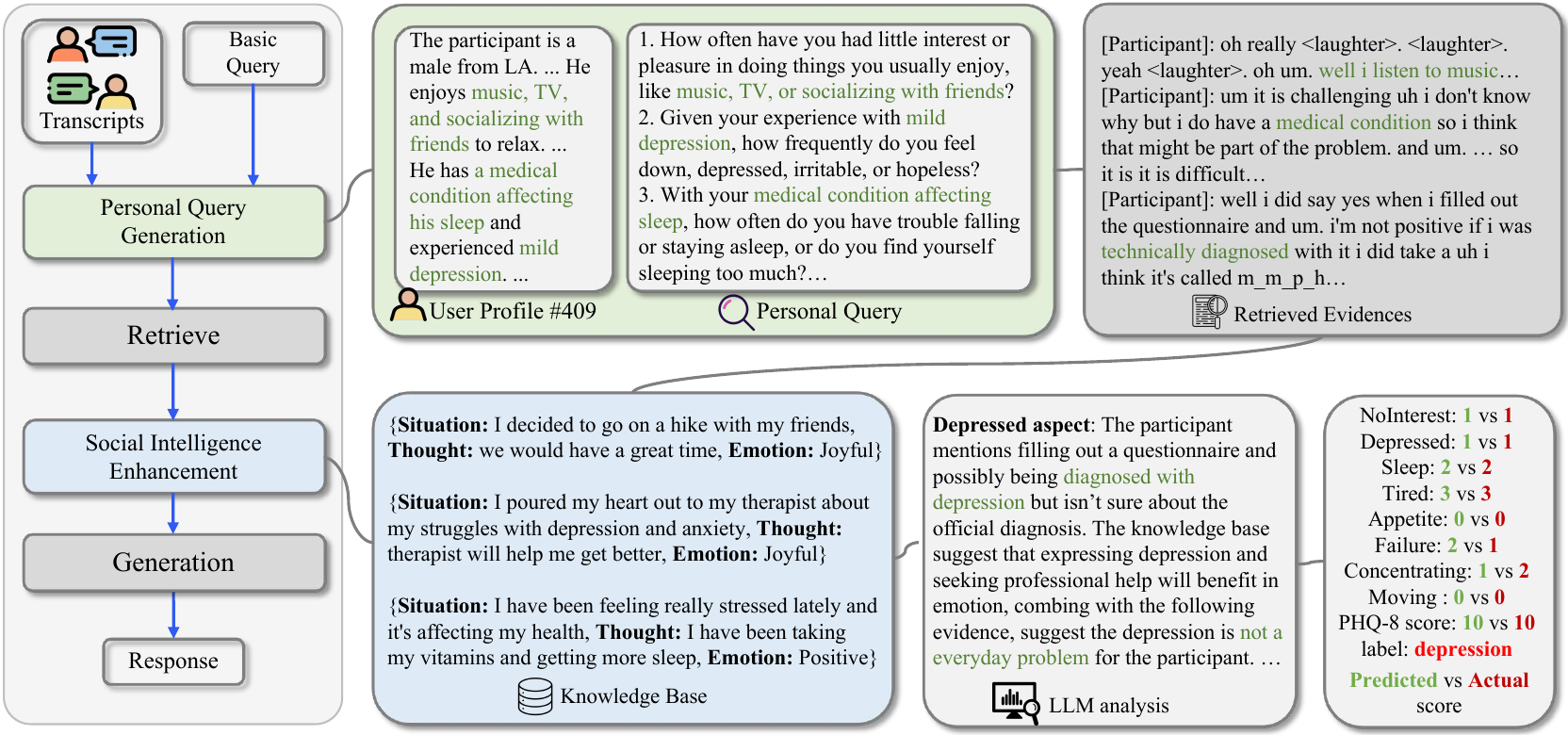}
    \caption{Case study for user \#409. Texts containing personal identification information are removed. Texts in \textcolor{green}{green} indicate the important information for prediction, and texts in \textcolor{red}{red} indicate the actual scores.}
    \label{fig:case_study}
\end{figure*}

\subsection{Ablation Studies}
To answer \textbf{RQ2}, we compare RED with its variants on the combined DAIC-WoZ set, merging the training and development sets. We focus on two subquestions: (1) How does the capability of backbone LLMs affect RED's performance? (2) How does the retrieval module setting impact performance? The results in Table~\ref{tab:full} reveal the following:

\paragraph{Training set is generally more difficult than the development set.}
As shown in Table~\ref{tab:full}, metrics for all variants on the combined set are lower than on the development set, suggesting the training set is more challenging. This is due to two factors: (1) The training data has a less clear decision boundary, with most participants' scores around 10, near the depression threshold, compared to the more extreme scores in the development set. (2) The training set contains more missing aspects; for example, more depressed participants reported issues with appetite and movement difficulties, leading to significant data gaps.

\paragraph{RED benefits from the improved capability of backbone LLMs.} 
As seen in Table~\ref{tab:full}, performance improves consistently with the enhanced backbone LLMs (GPT-4 to GPT-4o-mini and GPT-4o) across all retrieval settings, highlighting the importance of LLMs' reasoning and instruction-following capabilities in depression detection.

\paragraph{The retrieval process brings universal improvement.}
Regardless of the backbone LLM, the retrieval process improves results at each stage. Variants based on naive RAG outperform those with direct prompting, and personal RAG variants outperform naive RAG, offering more precise retrieval tailored to user backgrounds.

\subsection{Explanation Extraction Analysis}

\begin{table}[]
\centering
\resizebox{0.95\columnwidth}{!}{
\begin{tabular}{cllll}
\hline
\multicolumn{2}{c}{Method}          & \multicolumn{1}{c}{Precision} & \multicolumn{1}{c}{Recall} & \multicolumn{1}{c}{F1} \\ \hline
Non           & Direcet Prompt      & 30.42                         & 59.45                      & 40.21                       \\
Retrieval     & In-context Learning & 47.12                         & 62.42                      & 56.31                       \\ \hline
              & k=4                 & 93.53 & 34.66  & 50.58                           \\
Naive         & k=6                 & 91.78 & 49.38  & 64.21                          \\
RAG           & k=8                 & 89.70 & 59.64  & 71.64                           \\
              & k=10                & 88.30 & 69.02  & 77.48                        \\ \hline
              & k=4                 & \textbf{90.62} & 49.54  & 64.06               \\
Personal      & k=6                 & 88.78 & 63.01  & 73.71                      \\
RAG           & k=8                 & 87.39 & 72.76  & 79.41                       \\
              & k=10                & 86.78 & \textbf{80.19}  & \textbf{83.35}      \\ \hline
\end{tabular}}
\caption{Performance of RED's variants on the evidence extraction on DAIC-WoZ benchmark. The best scores are in \textbf{bold}. }
\label{tab:evidence}
\end{table}

To answer \textbf{RQ3}, we compare the performance of RED with other baseline models on evidence extraction. The evidence, annotated by \cite{agarwal-etal-2024-analyzing}, consists of text chunks identified by human annotators as important for depression prediction. The results, shown in Table~\ref{tab:evidence}, demonstrate RED’s effectiveness in evidence extraction.
Notably, without the retrieval system, the precision of evidence retrieval is much lower compared to the retrieval-based system, indicating that LLMs tend to generate hallucinations. With the retrieval module, both precision and recall increase significantly, and these improvements are further enhanced with the personal retrieval module. This highlights the effectiveness of personal query generation, which tailors the retrieval process to the user’s background.

\subsection{Case Study}
To answer \textbf{RQ4}, we analyze a sample from the development set to demonstrate how RED works. Personal identification information is removed.
As shown in Figure~\ref{fig:case_study}, this participant has a PHQ-8 score of 10, placing him at the threshold of depression, making this a challenging case to predict. However, RED successfully predicted this case, with four out of the eight aspect scores matching the ground truth.
The score for the \textit{depressed} aspect is particularly difficult, as the participant had previously been diagnosed with depression. Many systems would have incorrectly predicted this aspect with a score of 3. However, with the knowledge base enhancement and the retrieved content, RED recognized that the participant had sought help, which greatly improved his condition, leading to a predicted score of 1.

\section{Conclusion}
In this paper, we introduced RED, a Retrieval-Augmented Generation framework designed for explainable depression detection. 
By retrieving evidence from clinical interview transcripts, RED not only provides transparent explanations for its predictions but also adapts to individual user contexts through personalized query generation.
Furthermore, we enhanced the RAG by retrieving social intelligence knowledge with an event-centric retriever. 
Experimental results on the real-world dataset validate the effectiveness of RED, demonstrating its effectiveness in explainable depression detection.

\section*{Limitations}
We identify three key limitations in RED. 

\textbf{1) Corpus Size.}
Due to data collection challenges and privacy concerns, datasets for depression detection in clinical interviews are often limited in size. In our experiments, we report the average performance across multiple runs and conduct significance testing to ensure that any observed improvements are statistically valid. However, the small size of the datasets may still impact the generalizability of the results.

\textbf{2) Single Modality.}
The DAIC dataset includes multiple modalities, such as audio and video, which could potentially enhance depression detection. Our proposed method focuses primarily on text, which is considered the most informative and widely used modality for this task. Additionally, text serves as the safest modality for protecting user privacy. That said, future work should explore the potential for explainable depression detection using multimodal data, integrating other modalities to improve the system's performance and robustness.

\textbf{3) Task Format.}
This work focuses on explainable depression detection within clinical interviews, where dialogue snippets provide the evidence for judgment through a Retrieval-augmented Generation (RAG) framework. This setting requires interviews to be of sufficient length to provide adequate evidence for retrieval. As a result, we were unable to experiment with the EATD dataset, where each participant responded to only three questions. Additionally, the proposed method may not be easily transferred to other important settings, such as detecting early signs of depression from social media posts, due to significant differences in data structure, task format, and judgment criteria.

\section*{Ethical Impact}
In developing RED for explainable depression detection, we acknowledge several potential ethical concerns related to data privacy, fairness, the role of AI in mental health, and responsible use of datasets.

1) \textbf{Data Privacy and Consent}
RED utilizes clinical interview transcripts to detect depression. These datasets are crucial for building and testing the model, and we emphasize the importance of obtaining informed consent from all participants. Any personal information, such as names or identifying details, must be anonymized or removed before use to ensure privacy. Given that mental health data is particularly sensitive, stringent privacy safeguards, such as data encryption and secure handling, must be in place to protect participants from any unintended disclosures.

2) \textbf{Bias and Fairness}
While RED tailors its predictions using personalized query generation based on user background, it is essential to ensure that the model does not inadvertently introduce or amplify biases. The data used for training and testing must be representative of diverse populations to avoid reinforcing stereotypes or underrepresenting specific groups, particularly those who may be vulnerable to mental health issues. We must carefully monitor the model's outputs to ensure fairness and continuous efforts should be made to detect and mitigate any bias in the system, particularly regarding sensitive demographic factors such as age, gender, or ethnicity.

3) \textbf{Role of AI in Mental Health Diagnosis}  
The use of RED in mental health settings should always complement, not replace, clinical expertise. While the system aims to provide valuable insights and explanations through explainable predictions, the final diagnosis and treatment decisions should remain the responsibility of qualified healthcare professionals. Using RED as an automated system for diagnosis without human oversight could lead to the misinterpretation of results, potentially harming users. The model's outputs should be viewed as recommendations or support tools, with the understanding that human judgment is essential for accurate mental health care.

4) \textbf{Responsible Dataset Use and Access}  
The datasets used for training RED must be handled responsibly and in compliance with all relevant ethical standards. All data must be obtained with the appropriate permissions and used strictly for research purposes. We must adhere to institutional and legal requirements when accessing and utilizing these datasets, ensuring they are not shared or disseminated without proper authorization. Further, when working with clinical datasets, it is critical to respect participant confidentiality and uphold ethical standards in all stages of data usage.

By addressing these ethical concerns, we can ensure that RED is developed and deployed in a responsible, transparent, and equitable manner, prioritizing user well-being and promoting trust in AI-driven mental health tools.

\section*{Acknowledgments}
This work was supported by the UK Engineering and Physical Sciences Research Council (EPSRC) through a Turing AI Fellowship (grant no. EP/V020579/1, EP/V020579/2). 
The work was also supported by the National Natural Science Foundation of China (62176053).


\begin{thebibliography}{50}
\providecommand{\natexlab}[1]{#1}

\bibitem[{Agarwal et~al.(2024)Agarwal, Milintsevich, Metivier, Rotharmel, Dias, and Dollfus}]{agarwal-etal-2024-analyzing}
Navneet Agarwal, Kirill Milintsevich, Lucie Metivier, Maud Rotharmel, Ga{\"e}l Dias, and Sonia Dollfus. 2024.
\newblock \href {https://aclanthology.org/2024.lrec-main.87/} {Analyzing symptom-based depression level estimation through the prism of psychiatric expertise}.
\newblock In \emph{Proceedings of the 2024 Joint International Conference on Computational Linguistics, Language Resources and Evaluation (LREC-COLING 2024)}, pages 974--983, Torino, Italia. ELRA and ICCL.

\bibitem[{Al~Hanai et~al.(2018)Al~Hanai, Ghassemi, and Glass}]{al2018detecting}
Tuka Al~Hanai, Mohammad~M Ghassemi, and James~R Glass. 2018.
\newblock Detecting depression with audio/text sequence modeling of interviews.
\newblock In \emph{Interspeech}, pages 1716--1720.

\bibitem[{An et~al.(2020)An, Wang, Li, and Zhou}]{an2020multimodal}
Minghui An, Jingjing Wang, Shoushan Li, and Guodong Zhou. 2020.
\newblock Multimodal topic-enriched auxiliary learning for depression detection.
\newblock In \emph{proceedings of the 28th international conference on computational linguistics}, pages 1078--1089.

\bibitem[{Asai et~al.(2023)Asai, Wu, Wang, Sil, and Hajishirzi}]{asai2023self}
Akari Asai, Zeqiu Wu, Yizhong Wang, Avirup Sil, and Hannaneh Hajishirzi. 2023.
\newblock Self-rag: Learning to retrieve, generate, and critique through self-reflection.
\newblock \emph{arXiv preprint arXiv:2310.11511}.

\bibitem[{Burdisso et~al.(2023)Burdisso, Villatoro-Tello, Madikeri, and Motlicek}]{burdisso2023node}
Sergio Burdisso, Esa{\'u} Villatoro-Tello, Srikanth Madikeri, and Petr Motlicek. 2023.
\newblock Node-weighted graph convolutional network for depression detection in transcribed clinical interviews.
\newblock \emph{arXiv preprint arXiv:2307.00920}.

\bibitem[{Burdisso et~al.(2019{\natexlab{a}})Burdisso, Errecalde, and Montes-y G{\'o}mez}]{burdisso2019text}
Sergio~G Burdisso, Marcelo Errecalde, and Manuel Montes-y G{\'o}mez. 2019{\natexlab{a}}.
\newblock A text classification framework for simple and effective early depression detection over social media streams.
\newblock \emph{Expert Systems with Applications}, 133:182--197.

\bibitem[{Burdisso et~al.(2020)Burdisso, Errecalde, and Montes-y G{\'o}mez}]{burdisso2020tau}
Sergio~G Burdisso, Marcelo Errecalde, and Manuel Montes-y G{\'o}mez. 2020.
\newblock $\tau$-ss3: A text classifier with dynamic n-grams for early risk detection over text streams.
\newblock \emph{Pattern Recognition Letters}, 138:130--137.

\bibitem[{Burdisso et~al.(2019{\natexlab{b}})Burdisso, Errecalde, and Montes-y G{\'o}mez}]{burdisso2019unsl}
Sergio~Gast{\'o}n Burdisso, Marcelo Errecalde, and Manuel Montes-y G{\'o}mez. 2019{\natexlab{b}}.
\newblock Unsl at erisk 2019: a unified approach for anorexia, self-harm and depression detection in social media.
\newblock In \emph{CLEF (Working Notes)}.

\bibitem[{Chen et~al.(2024)Chen, Deng, Zhou, Wu, Qian, and Huang}]{chen-etal-2024-depression}
Zhuang Chen, Jiawen Deng, Jinfeng Zhou, Jincenzi Wu, Tieyun Qian, and Minlie Huang. 2024.
\newblock \href {https://doi.org/10.18653/v1/2024.naacl-long.452} {Depression detection in clinical interviews with {LLM}-empowered structural element graph}.
\newblock In \emph{Proceedings of the 2024 Conference of the North American Chapter of the Association for Computational Linguistics: Human Language Technologies (Volume 1: Long Papers)}, pages 8181--8194, Mexico City, Mexico. Association for Computational Linguistics.

\bibitem[{Fang et~al.(2023)Fang, Peng, Liang, Hung, and Liu}]{FANG2023104561}
Ming Fang, Siyu Peng, Yujia Liang, Chih-Cheng Hung, and Shuhua Liu. 2023.
\newblock \href {https://doi.org/10.1016/j.bspc.2022.104561} {A multimodal fusion model with multi-level attention mechanism for depression detection}.
\newblock \emph{Biomedical Signal Processing and Control}, 82:104561.

\bibitem[{Fava and Kendler(2000)}]{fava2000major}
Maurizio Fava and Kenneth~S Kendler. 2000.
\newblock Major depressive disorder.
\newblock \emph{Neuron}, 28(2):335--341.

\bibitem[{Gao et~al.(2023)Gao, Xiong, Gao, Jia, Pan, Bi, Dai, Sun, and Wang}]{gao2023retrieval}
Yunfan Gao, Yun Xiong, Xinyu Gao, Kangxiang Jia, Jinliu Pan, Yuxi Bi, Yi~Dai, Jiawei Sun, and Haofen Wang. 2023.
\newblock Retrieval-augmented generation for large language models: A survey.
\newblock \emph{arXiv preprint arXiv:2312.10997}.

\bibitem[{Goldman et~al.(1999)Goldman, Nielsen, Champion, and Council~on Scientific~Affairs}]{goldman1999awareness}
Larry~S Goldman, Nancy~H Nielsen, Hunter~C Champion, and American Medical~Association Council~on Scientific~Affairs. 1999.
\newblock Awareness, diagnosis, and treatment of depression.
\newblock \emph{Journal of general internal medicine}, 14(9):569--580.

\bibitem[{Gratch et~al.(2014)Gratch, Artstein, Lucas, Stratou, Scherer, Nazarian, Wood, Boberg, DeVault, Marsella et~al.}]{gratch2014distress}
Jonathan Gratch, Ron Artstein, Gale~M Lucas, Giota Stratou, Stefan Scherer, Angela Nazarian, Rachel Wood, Jill Boberg, David DeVault, Stacy Marsella, et~al. 2014.
\newblock The distress analysis interview corpus of human and computer interviews.
\newblock In \emph{LREC}, volume~14, pages 3123--3128. Reykjavik.

\bibitem[{Guu et~al.(2020)Guu, Lee, Tung, Pasupat, and Chang}]{guu2020retrieval}
Kelvin Guu, Kenton Lee, Zora Tung, Panupong Pasupat, and Mingwei Chang. 2020.
\newblock Retrieval augmented language model pre-training.
\newblock In \emph{International conference on machine learning}, pages 3929--3938. PMLR.

\bibitem[{Hou et~al.(2024)Hou, Zhang, Shen, Tan, Shen, and Lu}]{hou2024entering}
Guiyang Hou, Wenqi Zhang, Yongliang Shen, Zeqi Tan, Sihao Shen, and Weiming Lu. 2024.
\newblock Entering real social world! benchmarking the theory of mind and socialization capabilities of llms from a first-person perspective.
\newblock \emph{arXiv preprint arXiv:2410.06195}.

\bibitem[{Islam et~al.(2018)Islam, Kabir, Ahmed, Kamal, Wang, and Ulhaq}]{islam2018depression}
Md~Rafiqul Islam, Muhammad~Ashad Kabir, Ashir Ahmed, Abu Raihan~M Kamal, Hua Wang, and Anwaar Ulhaq. 2018.
\newblock Depression detection from social network data using machine learning techniques.
\newblock \emph{Health information science and systems}, 6:1--12.

\bibitem[{Izacard et~al.(2022)Izacard, Lewis, Lomeli, Hosseini, Petroni, Schick, Dwivedi-Yu, Joulin, Riedel, and Grave}]{izacard2022few}
Gautier Izacard, Patrick Lewis, Maria Lomeli, Lucas Hosseini, Fabio Petroni, Timo Schick, Jane Dwivedi-Yu, Armand Joulin, Sebastian Riedel, and Edouard Grave. 2022.
\newblock Few-shot learning with retrieval augmented language models.
\newblock \emph{arXiv preprint arXiv:2208.03299}, 1(2):4.

\bibitem[{Jiang et~al.(2023)Jiang, Xu, Gao, Sun, Liu, Dwivedi-Yu, Yang, Callan, and Neubig}]{jiang-etal-2023-active}
Zhengbao Jiang, Frank Xu, Luyu Gao, Zhiqing Sun, Qian Liu, Jane Dwivedi-Yu, Yiming Yang, Jamie Callan, and Graham Neubig. 2023.
\newblock \href {https://doi.org/10.18653/v1/2023.emnlp-main.495} {Active retrieval augmented generation}.
\newblock In \emph{Proceedings of the 2023 Conference on Empirical Methods in Natural Language Processing}, pages 7969--7992, Singapore. Association for Computational Linguistics.

\bibitem[{Kroenke et~al.(2001)Kroenke, Spitzer, and Williams}]{kroenke2001phq}
Kurt Kroenke, Robert~L Spitzer, and Janet~BW Williams. 2001.
\newblock The phq-9: validity of a brief depression severity measure.
\newblock \emph{Journal of general internal medicine}, 16(9):606--613.

\bibitem[{Kroenke et~al.(2009)Kroenke, Strine, Spitzer, Williams, Berry, and Mokdad}]{kroenke2009phq}
Kurt Kroenke, Tara~W Strine, Robert~L Spitzer, Janet~BW Williams, Joyce~T Berry, and Ali~H Mokdad. 2009.
\newblock The phq-8 as a measure of current depression in the general population.
\newblock \emph{Journal of affective disorders}, 114(1-3):163--173.

\bibitem[{Lewis et~al.(2020)Lewis, Perez, Piktus, Petroni, Karpukhin, Goyal, K{\"u}ttler, Lewis, Yih, Rockt{\"a}schel et~al.}]{lewis2020retrieval}
Patrick Lewis, Ethan Perez, Aleksandra Piktus, Fabio Petroni, Vladimir Karpukhin, Naman Goyal, Heinrich K{\"u}ttler, Mike Lewis, Wen-tau Yih, Tim Rockt{\"a}schel, et~al. 2020.
\newblock Retrieval-augmented generation for knowledge-intensive nlp tasks.
\newblock \emph{Advances in Neural Information Processing Systems}, 33:9459--9474.

\bibitem[{Lin et~al.(2024)Lin, Chen, Chen, Shi, Lomeli, James, Rodriguez, Kahn, Szilvasy, Lewis, Zettlemoyer, and tau Yih}]{lin2024radit}
Xi~Victoria Lin, Xilun Chen, Mingda Chen, Weijia Shi, Maria Lomeli, Richard James, Pedro Rodriguez, Jacob Kahn, Gergely Szilvasy, Mike Lewis, Luke Zettlemoyer, and Wen tau Yih. 2024.
\newblock \href {https://openreview.net/forum?id=22OTbutug9} {{RA}-{DIT}: Retrieval-augmented dual instruction tuning}.
\newblock In \emph{The Twelfth International Conference on Learning Representations}.

\bibitem[{Liu et~al.(2024)Liu, Anand, Zhou, Huang, and Zhao}]{liu-etal-2024-interintent}
Ziyi Liu, Abhishek Anand, Pei Zhou, Jen-tse Huang, and Jieyu Zhao. 2024.
\newblock \href {https://doi.org/10.18653/v1/2024.emnlp-main.383} {{I}nter{I}ntent: Investigating social intelligence of {LLM}s via intention understanding in an interactive game context}.
\newblock In \emph{Proceedings of the 2024 Conference on Empirical Methods in Natural Language Processing}, pages 6718--6746, Miami, Florida, USA. Association for Computational Linguistics.

\bibitem[{Luo et~al.(2023)Luo, Chuang, Gong, Zhang, Kim, Wu, Fox, Meng, and Glass}]{luo2023sail}
Hongyin Luo, Yung-Sung Chuang, Yuan Gong, Tianhua Zhang, Yoon Kim, Xixin Wu, Danny Fox, Helen Meng, and James Glass. 2023.
\newblock Sail: Search-augmented instruction learning.
\newblock \emph{arXiv preprint arXiv:2305.15225}.

\bibitem[{Ma et~al.(2016)Ma, Yang, Chen, Huang, and Wang}]{ma2016depaudionet}
Xingchen Ma, Hongyu Yang, Qiang Chen, Di~Huang, and Yunhong Wang. 2016.
\newblock Depaudionet: An efficient deep model for audio based depression classification.
\newblock In \emph{Proceedings of the 6th international workshop on audio/visual emotion challenge}, pages 35--42.

\bibitem[{Mallen et~al.(2023)Mallen, Asai, Zhong, Das, Khashabi, and Hajishirzi}]{mallen-etal-2023-trust}
Alex Mallen, Akari Asai, Victor Zhong, Rajarshi Das, Daniel Khashabi, and Hannaneh Hajishirzi. 2023.
\newblock \href {https://doi.org/10.18653/v1/2023.acl-long.546} {When not to trust language models: Investigating effectiveness of parametric and non-parametric memories}.
\newblock In \emph{Proceedings of the 61st Annual Meeting of the Association for Computational Linguistics (Volume 1: Long Papers)}, pages 9802--9822, Toronto, Canada. Association for Computational Linguistics.

\bibitem[{Mallol-Ragolta et~al.(2019)Mallol-Ragolta, Zhao, Stappen, Cummins, and Schuller}]{mallol2019hierarchical}
Adria Mallol-Ragolta, Ziping Zhao, Lukas Stappen, Nicholas Cummins, and Bj{\"o}rn Schuller. 2019.
\newblock A hierarchical attention network-based approach for depression detection from transcribed clinical interviews.

\bibitem[{Niu et~al.(2021)Niu, Chen, Chen, and Yang}]{9413486}
Meng Niu, Kai Chen, Qingcai Chen, and Lufeng Yang. 2021.
\newblock \href {https://doi.org/10.1109/ICASSP39728.2021.9413486} {Hcag: A hierarchical context-aware graph attention model for depression detection}.
\newblock In \emph{ICASSP 2021 - 2021 IEEE International Conference on Acoustics, Speech and Signal Processing (ICASSP)}, pages 4235--4239.

\bibitem[{Orabi et~al.(2018)Orabi, Buddhitha, Orabi, and Inkpen}]{orabi2018deep}
Ahmed~Husseini Orabi, Prasadith Buddhitha, Mahmoud~Husseini Orabi, and Diana Inkpen. 2018.
\newblock Deep learning for depression detection of twitter users.
\newblock In \emph{Proceedings of the fifth workshop on computational linguistics and clinical psychology: from keyboard to clinic}, pages 88--97.

\bibitem[{Richardson et~al.(2023)Richardson, Zhang, Gillespie, Kar, Singh, Raeesy, Khan, and Sethy}]{richardson2023integrating}
Chris Richardson, Yao Zhang, Kellen Gillespie, Sudipta Kar, Arshdeep Singh, Zeynab Raeesy, Omar~Zia Khan, and Abhinav Sethy. 2023.
\newblock Integrating summarization and retrieval for enhanced personalization via large language models.
\newblock \emph{arXiv preprint arXiv:2310.20081}.

\bibitem[{Sadeque et~al.(2018)Sadeque, Xu, and Bethard}]{sadeque2018measuring}
Farig Sadeque, Dongfang Xu, and Steven Bethard. 2018.
\newblock Measuring the latency of depression detection in social media.
\newblock In \emph{Proceedings of the Eleventh ACM International Conference on Web Search and Data Mining}, pages 495--503.

\bibitem[{Salas-Z{\'a}rate et~al.(2022)Salas-Z{\'a}rate, Alor-Hern{\'a}ndez, Salas-Z{\'a}rate, Paredes-Valverde, Bustos-L{\'o}pez, and S{\'a}nchez-Cervantes}]{salas2022detecting}
Rafael Salas-Z{\'a}rate, Giner Alor-Hern{\'a}ndez, Mar{\'\i}a del~Pilar Salas-Z{\'a}rate, Mario~Andr{\'e}s Paredes-Valverde, Maritza Bustos-L{\'o}pez, and Jos{\'e}~Luis S{\'a}nchez-Cervantes. 2022.
\newblock Detecting depression signs on social media: a systematic literature review.
\newblock In \emph{Healthcare}, volume~10, page 291. MDPI.

\bibitem[{Sardari et~al.(2022)Sardari, Nakisa, Rastgoo, and Eklund}]{sardari2022audio}
Sara Sardari, Bahareh Nakisa, Mohammed~Naim Rastgoo, and Peter Eklund. 2022.
\newblock Audio based depression detection using convolutional autoencoder.
\newblock \emph{Expert Systems with Applications}, 189:116076.

\bibitem[{Schick et~al.(2023)Schick, Dwivedi-Yu, Dess{\`\i}, Raileanu, Lomeli, Hambro, Zettlemoyer, Cancedda, and Scialom}]{schick2023toolformer}
Timo Schick, Jane Dwivedi-Yu, Roberto Dess{\`\i}, Roberta Raileanu, Maria Lomeli, Eric Hambro, Luke Zettlemoyer, Nicola Cancedda, and Thomas Scialom. 2023.
\newblock Toolformer: Language models can teach themselves to use tools.
\newblock \emph{Advances in Neural Information Processing Systems}, 36:68539--68551.

\bibitem[{Shen et~al.(2022{\natexlab{a}})Shen, Yang, and Lin}]{shen2022automatic}
Ying Shen, Huiyu Yang, and Lin Lin. 2022{\natexlab{a}}.
\newblock Automatic depression detection: An emotional audio-textual corpus and a gru/bilstm-based model.
\newblock In \emph{ICASSP 2022-2022 IEEE International Conference on Acoustics, Speech and Signal Processing (ICASSP)}, pages 6247--6251. IEEE.

\bibitem[{Shen et~al.(2022{\natexlab{b}})Shen, Yang, and Lin}]{9746569}
Ying Shen, Huiyu Yang, and Lin Lin. 2022{\natexlab{b}}.
\newblock \href {https://doi.org/10.1109/ICASSP43922.2022.9746569} {Automatic depression detection: an emotional audio-textual corpus and a gru/bilstm-based model}.
\newblock In \emph{ICASSP 2022 - 2022 IEEE International Conference on Acoustics, Speech and Signal Processing (ICASSP)}, pages 6247--6251.

\bibitem[{Shi et~al.(2023)Shi, Chen, Misra, Scales, Dohan, Chi, Sch{\"a}rli, and Zhou}]{shi2023large}
Freda Shi, Xinyun Chen, Kanishka Misra, Nathan Scales, David Dohan, Ed~H Chi, Nathanael Sch{\"a}rli, and Denny Zhou. 2023.
\newblock Large language models can be easily distracted by irrelevant context.
\newblock In \emph{International Conference on Machine Learning}, pages 31210--31227. PMLR.

\bibitem[{Villatoro-Tello et~al.(2021)Villatoro-Tello, Ram{\'\i}rez-de-la Rosa, G{\'a}tica-P{\'e}rez, Magimai.-Doss, and Jim{\'e}nez-Salazar}]{villatoro2021approximating}
Esa{\'u} Villatoro-Tello, Gabriela Ram{\'\i}rez-de-la Rosa, Daniel G{\'a}tica-P{\'e}rez, Mathew Magimai.-Doss, and H{\'e}ctor Jim{\'e}nez-Salazar. 2021.
\newblock Approximating the mental lexicon from clinical interviews as a support tool for depression detection.
\newblock In \emph{Proceedings of the 2021 international conference on multimodal interaction}, pages 557--566.

\bibitem[{Wang et~al.(2023)Wang, Li, Yin, Wu, and Liu}]{wang2023emotional}
Xuena Wang, Xueting Li, Zi~Yin, Yue Wu, and Jia Liu. 2023.
\newblock Emotional intelligence of large language models.
\newblock \emph{Journal of Pacific Rim Psychology}, 17:18344909231213958.

\bibitem[{Wang et~al.(2024)Wang, Inkpen, and Kirinde~Gamaarachchige}]{wang-etal-2024-explainable}
Yuxi Wang, Diana Inkpen, and Prasadith Kirinde~Gamaarachchige. 2024.
\newblock \href {https://aclanthology.org/2024.clpsych-1.8/} {Explainable depression detection using large language models on social media data}.
\newblock In \emph{Proceedings of the 9th Workshop on Computational Linguistics and Clinical Psychology (CLPsych 2024)}, pages 108--126, St. Julians, Malta. Association for Computational Linguistics.

\bibitem[{Wu et~al.(2024)Wu, Chen, Deng, Sabour, Meng, and Huang}]{wu-etal-2024-coke}
Jincenzi Wu, Zhuang Chen, Jiawen Deng, Sahand Sabour, Helen Meng, and Minlie Huang. 2024.
\newblock \href {https://doi.org/10.18653/v1/2024.acl-long.848} {{COKE}: A cognitive knowledge graph for machine theory of mind}.
\newblock In \emph{Proceedings of the 62nd Annual Meeting of the Association for Computational Linguistics (Volume 1: Long Papers)}, pages 15984--16007, Bangkok, Thailand. Association for Computational Linguistics.

\bibitem[{Wu et~al.(2022)Wu, Wu, and Yu}]{wu2022climate}
Wen Wu, Mengyue Wu, and Kai Yu. 2022.
\newblock Climate and weather: Inspecting depression detection via emotion recognition.
\newblock In \emph{ICASSP 2022-2022 IEEE International Conference on Acoustics, Speech and Signal Processing (ICASSP)}, pages 6262--6266. IEEE.

\bibitem[{Xezonaki et~al.(2020)Xezonaki, Paraskevopoulos, Potamianos, and Narayanan}]{xezonaki2020affective}
Danai Xezonaki, Georgios Paraskevopoulos, Alexandros Potamianos, and Shrikanth Narayanan. 2020.
\newblock Affective conditioning on hierarchical networks applied to depression detection from transcribed clinical interviews.
\newblock \emph{arXiv preprint arXiv:2006.08336}.

\bibitem[{Yoon et~al.(2022)Yoon, Kang, Kim, and Han}]{yoon2022d}
Jeewoo Yoon, Chaewon Kang, Seungbae Kim, and Jinyoung Han. 2022.
\newblock D-vlog: Multimodal vlog dataset for depression detection.
\newblock In \emph{Proceedings of the AAAI Conference on Artificial Intelligence}, volume~36, pages 12226--12234.

\bibitem[{Yoran et~al.(2024)Yoran, Wolfson, Ram, and Berant}]{yoran2024making}
Ori Yoran, Tomer Wolfson, Ori Ram, and Jonathan Berant. 2024.
\newblock \href {https://openreview.net/forum?id=ZS4m74kZpH} {Making retrieval-augmented language models robust to irrelevant context}.
\newblock In \emph{The Twelfth International Conference on Learning Representations}.

\bibitem[{Zhang et~al.(2023)Zhang, Zhang, and Zhou}]{zhang-etal-2023-multi}
Linhai Zhang, Congzhi Zhang, and Deyu Zhou. 2023.
\newblock \href {https://doi.org/10.18653/v1/2023.findings-acl.384} {Multi-relational probabilistic event representation learning via projected {G}aussian embedding}.
\newblock In \emph{Findings of the Association for Computational Linguistics: ACL 2023}, pages 6162--6174, Toronto, Canada. Association for Computational Linguistics.

\bibitem[{Zhou et~al.(2023)Zhou, Yan, Shlapentokh-Rothman, Wang, and Wang}]{zhou2023language}
Andy Zhou, Kai Yan, Michal Shlapentokh-Rothman, Haohan Wang, and Yu-Xiong Wang. 2023.
\newblock Language agent tree search unifies reasoning acting and planning in language models.
\newblock \emph{arXiv preprint arXiv:2310.04406}.

\bibitem[{Zhou et~al.(2021)Zhou, Wang, Zhang, and He}]{zhou-etal-2021-implicit}
Deyu Zhou, Jianan Wang, Linhai Zhang, and Yulan He. 2021.
\newblock \href {https://doi.org/10.18653/v1/2021.emnlp-main.551} {Implicit sentiment analysis with event-centered text representation}.
\newblock In \emph{Proceedings of the 2021 Conference on Empirical Methods in Natural Language Processing}, pages 6884--6893, Online and Punta Cana, Dominican Republic. Association for Computational Linguistics.

\bibitem[{Zogan et~al.(2022)Zogan, Razzak, Wang, Jameel, and Xu}]{zogan2022explainable}
Hamad Zogan, Imran Razzak, Xianzhi Wang, Shoaib Jameel, and Guandong Xu. 2022.
\newblock Explainable depression detection with multi-aspect features using a hybrid deep learning model on social media.
\newblock \emph{World Wide Web}, 25(1):281--304.

\end{thebibliography}

\appendix
\section{Appendix}

\subsection{Prompt Template}
\label{app:prompt}

\begin{figure*}[]
    \centering
    \includegraphics[width=0.7\textwidth]{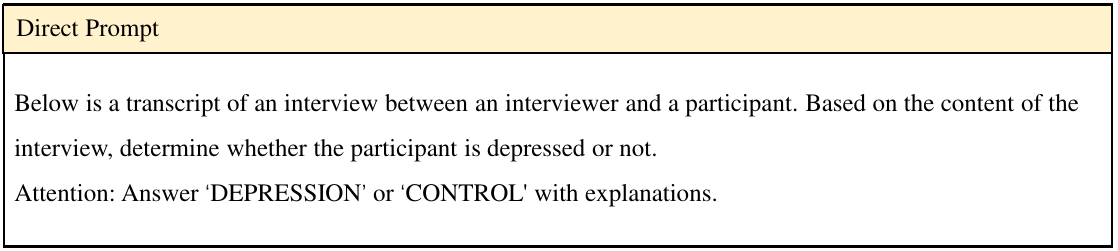}
    \caption{Prompt Template for Direct Prompting}
    \label{fig:direct_prompt}
\end{figure*}

\begin{figure*}[]
    \centering
    \includegraphics[width=1.0\textwidth]{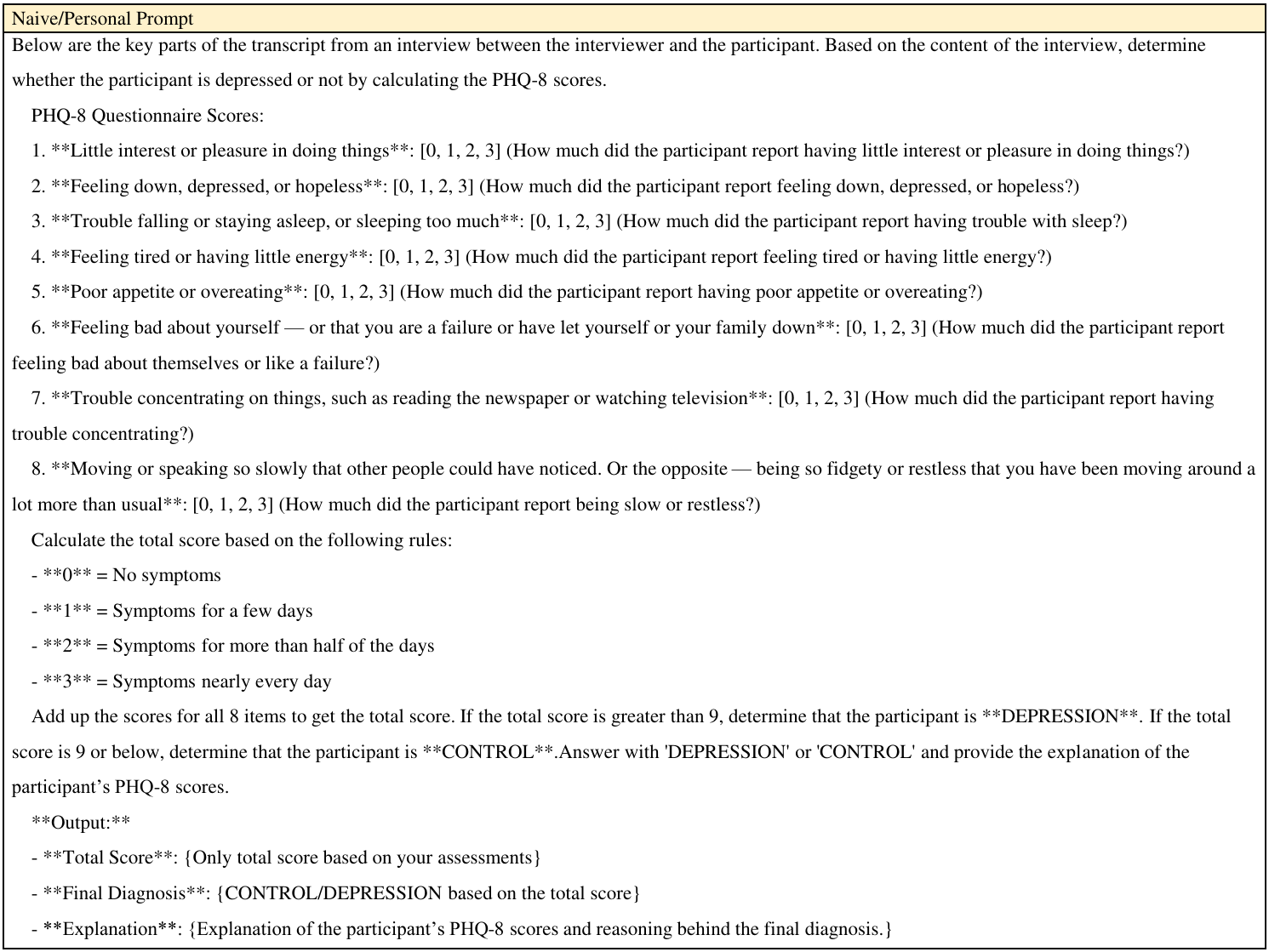}
    \caption{Prompt Template for Naive/Personal Prompt Retrieval}
    \label{fig:rag_prompt}
\end{figure*}

\begin{figure*}[]
    \centering
    \includegraphics[width=1.0\textwidth]{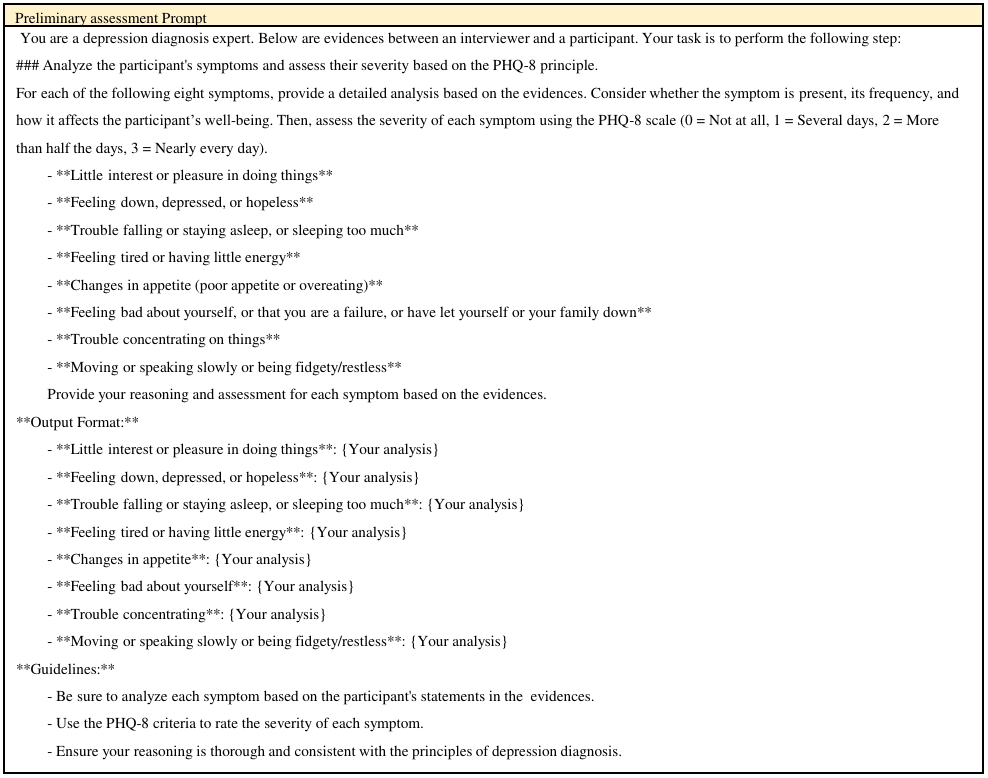}
    \caption{Prompt Template for Preliminary Assessment}
    \label{fig:preliminary_assessment}
\end{figure*}

\begin{figure*}[]
    \centering
    \includegraphics[width=1.0\textwidth]{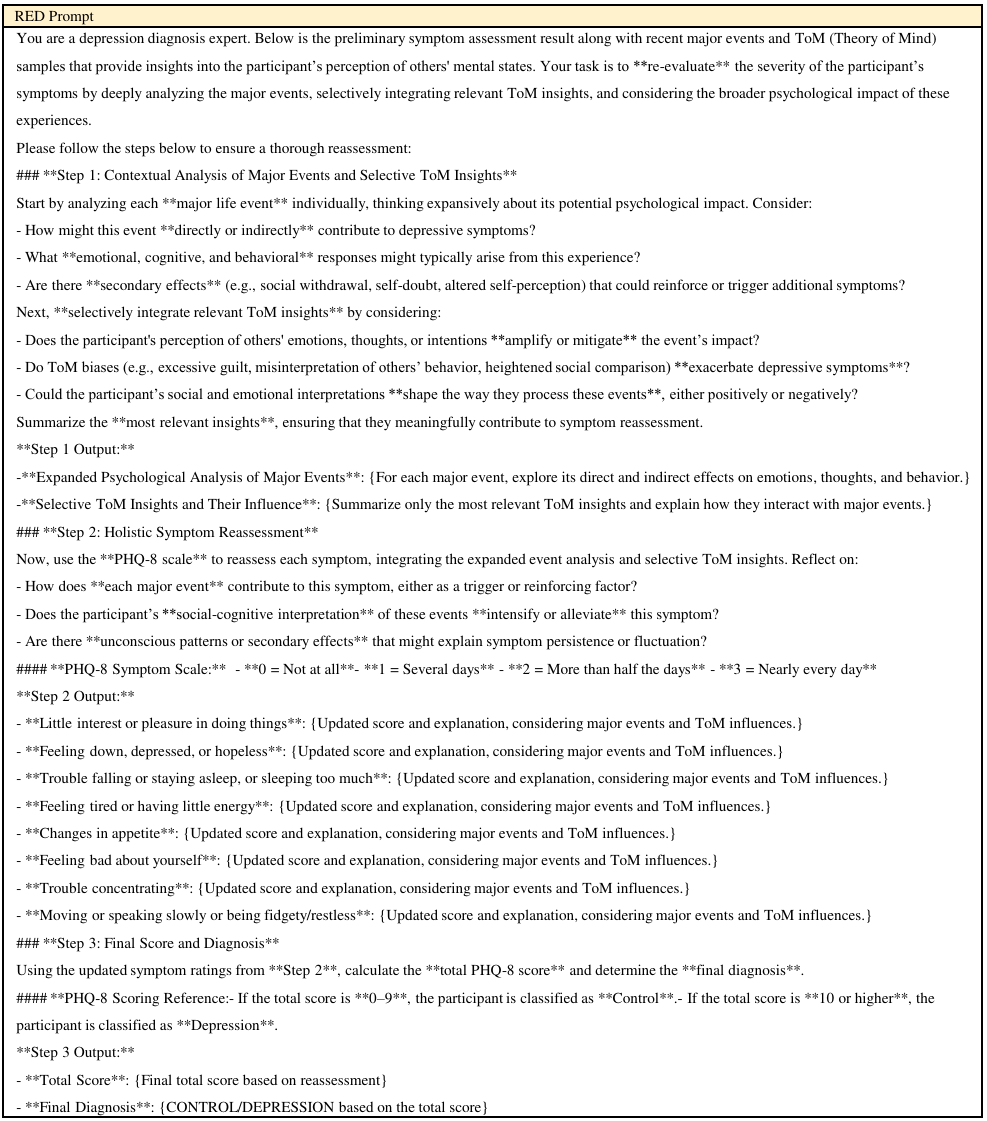}
    \caption{Prompt Template for RED Final Assessment}
    \label{fig:red_prompt}
\end{figure*}

\begin{figure*}[]
    \centering
    \includegraphics[width=1.0\textwidth]{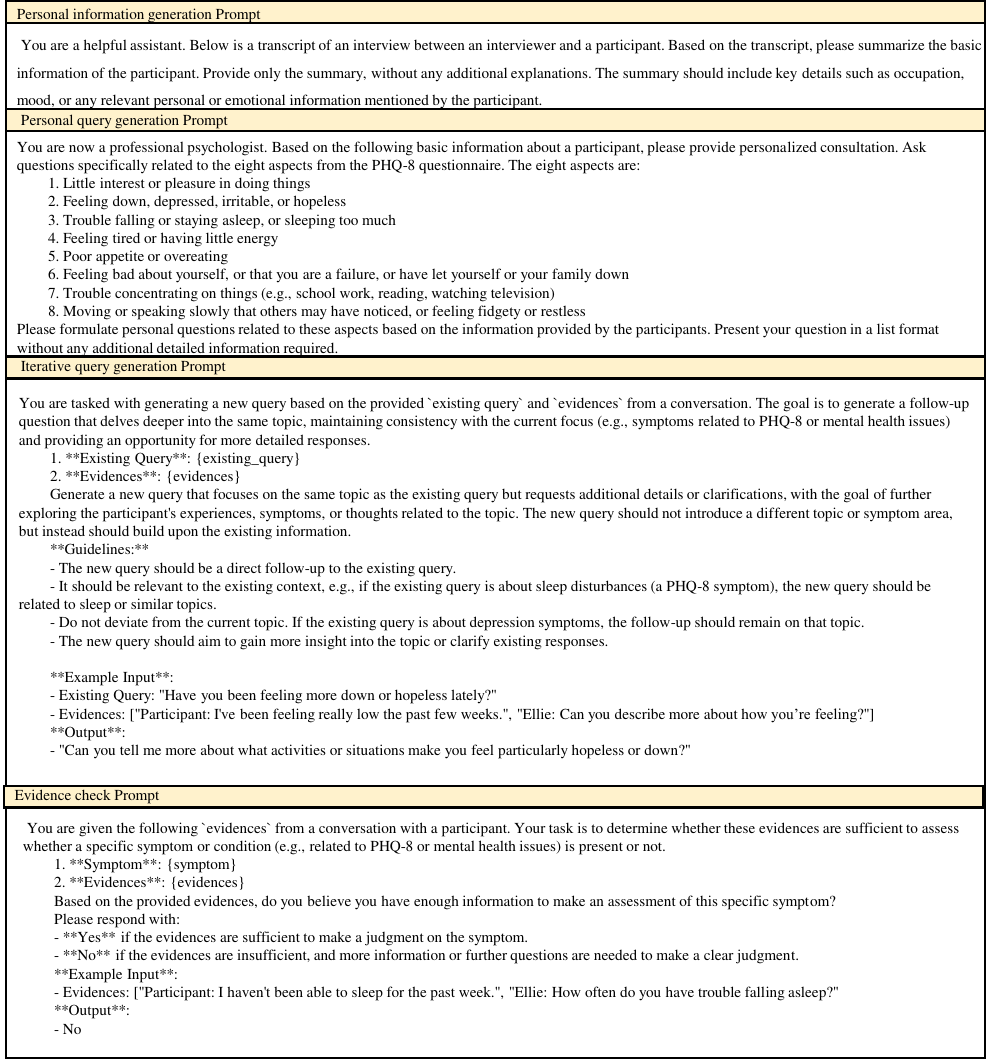}
    \caption{Prompt Template for Query Generation}
    \label{fig:query_generation}
\end{figure*}

\begin{figure*}[t]
    \centering
    \includegraphics[width=1.0\textwidth]{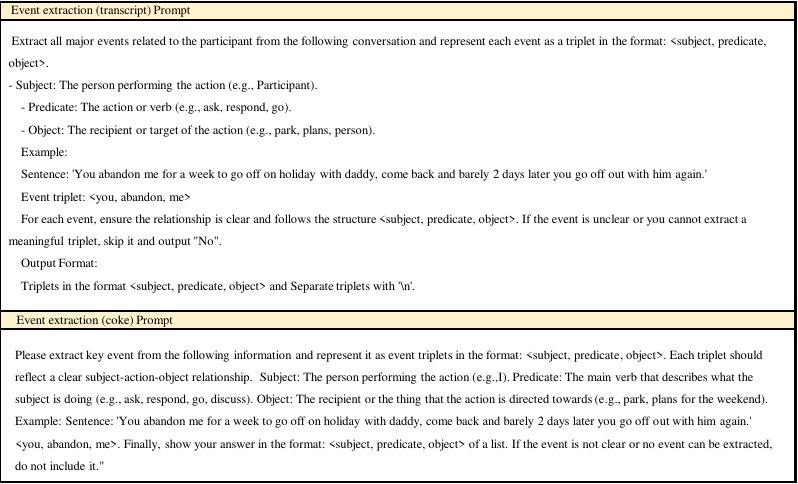}
    \caption{Prompt Template for Event Extraction}
    \label{fig:event_extraction}
\end{figure*}

In this section, we present all the prompt templates employed for LLM-based methods in our experiments. The prompt for direct prompting is shown in Figure~\ref{fig:direct_prompt}. 
The full prompt for personal query generation along with the basic query is shown in Figure~\ref{fig:query_generation}.
The Naive/Personal retrieval shares the same prompt, shown in Figure~\ref{fig:rag_prompt} as they are only different in the input queries.
The preliminary assessment Prompt which is employed between the retrieval module and social intelligence enhancement module is shown in Figure~\ref{fig:preliminary_assessment}. 
The event extraction prompt is shown in Figure~\ref{fig:event_extraction}.
The full prompt for RED after the social intelligence enhancement module is shown in Figure~\ref{fig:red_prompt}.

\subsection{Dataset Details}
\label{app:dataset}
In this section, we provide an overview of the raw data included in the dataset, focusing solely on the Transcript. The corpus includes full textual transcripts of each interview, capturing both the interviewer's questions and the participant's responses. Detailed statistics can be found in Table~\ref{tab:daic}.

\begin{table}[h]
\centering
\resizebox{0.95\columnwidth}{!}{
\begin{tabular}{cllll}
\hline
\multicolumn{1}{c}{Dataset} &\multicolumn{1}{c}{Size} & \multicolumn{1}{c}{Category} & \multicolumn{1}{c}{Round} 
 &\multicolumn{1}{c}{Token}\\ \hline
\multirow{2}{*}{Train} &\multirow{2}{*}{107} & [Deprssion]  30  &6,069  & 149,149    \\ 
                        &     & [Control]   77 & $\bar{x}=57$   & $\bar{x}=1,394$ \\ \hline
\multirow{2}{*}{Dev}  &\multirow{2}{*}{35}   & [Deprssion] 12 &1,909  & 53,588  \\ 
                       &             & [Control] 23 & $\bar{x}=55$    & $\bar{x}=1,531$      \\ \hline

\end{tabular}}
\caption{Detailed statistics of DAIC-WoZ.}
\label{tab:daic}
\end{table}

The COKE benchmark is a cognitive knowledge graph for machine theory of mind~\cite{wu-etal-2024-coke}. COKE contains a series of cognitive chains to describe human mental activities and behavioral/affective responses in social situations. It contains 5 dimensions, which are situation, thought, clue, action, and emotion. Detailed statistics can be found in Table~\ref{tab:coke}.

\begin{table}[h]
\centering
\resizebox{0.6\columnwidth}{!}{
\begin{tabular}{lcc}
\hline
\multicolumn{1}{c}{Dimension} & Count  & Avg. Len. \\ \hline
Siutation                     & 1,200  & 11.5      \\
Thought                       & 9,788  & 6.6       \\
Clue                          & 21,677 & 7.3       \\
Action                        & 19,875 & 6.8       \\
Emotion                       & 9,788  & 1.0       \\ \hline
\end{tabular}}
\caption{Detailed statistics of COKE.}
\label{tab:coke}
\end{table}

\subsection{Implementation Details}
\label{app:hyperparameter}

Given the length of the clinical interviews, we have chosen a context window of 128,000 tokens. For our depression detection task, we currently use the following models: \texttt{gpt-4o-0806}, \texttt{gpt-4o-mini-2024-07-18}, and \texttt{gpt-4-0125-preview}.

In the evidence retrieval section, we divide the corpus into chunks of 500 tokens. Then, we use \texttt{text-embedding-3-large} embeddings to encode both the queries and transcripts, applying the L2 distance metric to retrieve the top \(K\) evidence (with\(K=10\) by default). In other experimental settings,\(K\)  can also be 4, 6, or 8.
For the social intelligence component, we consider two embedding methods. The first method is the same as the one used in evidence retrieval, while the second uses event triples, which are encoded using MORE-CL. Both methods retrieve the top \(M\) evidence (with\(M=2\) by default).

\end{document}